\documentclass[runningheads]{llncs}

\usepackage[T1]{fontenc}
\usepackage{graphicx,verbatim}
\usepackage{amsfonts}
\usepackage{amsmath}
\usepackage{booktabs}
\usepackage{makecell}
\usepackage{tabularx}
\newcolumntype{Y}{>{\centering\arraybackslash}X}
\usepackage{hyperref}
\usepackage{color}

\begin{document}
\title{SSL-AD: Spatiotemporal Self-Supervised Learning for Generalizability and Adaptability Across Alzheimer’s Prediction Tasks and Datasets}
\titlerunning{SSL-AD: Spatiotemporal Self-Supervised Learning for Alzheimer's Disease}

\author{Emily Kaczmarek \and
Justin Szeto \and
Brennan Nichyporuk \and Tal Arbel}
%
\authorrunning{E. Kaczmarek et al.}
\institute{Centre for Intelligent Machines, McGill University, Montreal, Canada \\
Mila -- Quebec AI Institute, Montreal, Canada \\
\email{emily.kaczmarek@mail.mcgill.ca}}

\maketitle
\begin{abstract}
Alzheimer's disease is a progressive, neurodegenerative disorder that causes memory loss and cognitive decline. While there has been extensive research in applying deep learning models to Alzheimer's prediction tasks, these models remain limited by lack of available labeled data, poor generalization across datasets, and inflexibility to varying numbers of input scans and time intervals between scans. In this study, we adapt three state-of-the-art temporal self-supervised learning (SSL) approaches for 3D brain MRI analysis, and add novel extensions designed to handle variable-length inputs and learn robust spatial features. We aggregate four publicly available datasets comprising 3,161 patients for pre-training, and show the performance of our model across multiple Alzheimer's prediction tasks including diagnosis classification, conversion detection, and future conversion prediction. Importantly, our SSL model implemented with temporal order prediction and contrastive learning outperforms supervised learning on six out of seven downstream tasks. It demonstrates adaptability and generalizability across tasks and number of input images with varying time intervals, highlighting its capacity for robust performance across clinical applications. We release our code and model publicly at \url{https://github.com/emilykaczmarek/SSL-AD}. 

\end{abstract}

\section{Introduction}
Many chronic and neurodegenerative disorders progress gradually over time, making longitudinal prediction an important but challenging clinical goal. Alzheimer's Disease (AD) is an incurable, neurodegenerative disorder characterized by progressive memory loss, cognitive decline, and behavioural changes \cite{admci}. Typically, AD develops gradually, with Cognitively Normal (CN) individuals first progressing to Mild Cognitive Impairment (MCI), and potentially further transitioning to AD. AD motivates a range of clinically important spatial and temporal prediction tasks, including (i) current disease status, (ii) detecting the transition to MCI/AD, and (iii) predicting future transition to MCI/AD. Deep learning has been extensively applied to these tasks, with multiple review papers summarizing the progress in this field \cite{sina,tory,Taeho,valizadeh}. The majority of existing studies use supervised learning, which requires labeled data and separate models trained for each specific task. This severely limits both the amount of training data available, and the generalizability across tasks and/or population changes. In addition, most approaches are constrained to the fixed number of images they were trained with, making them inflexible in clinical settings where patients have multiple visits that can be leveraged to improve predictions. While supervised learning has shown strong results for Alzheimer's prediction tasks, it remains limited by the availability of labeled data, poor generalizability of representations, and lack of adaptability to varying numbers of images acquired per patient.

Recently, in machine learning, self-supervised learning (SSL) has emerged as a popular alternative to supervised learning \cite{ericsson}. In SSL, general-purpose representations are learned by deriving tasks using the input data alone (i.e., without external labels). This allows models to be trained on larger, unlabeled datasets (often aggregated from many sources), and fine-tuned using limited labels on numerous downstream tasks. In addition to spatial SSL models, several temporal SSL approaches have been introduced, including temporal order verification (predicting if a sequence of images is in the correct order) \cite{shuffleandlearn} and temporal order prediction (predicting the exact permutation of images) \cite{vcop}. These models are well-suited to analyzing 3D brain MRI of neurodegenerative diseases; given the similarity of brain images across time, the model will be encouraged to focus on subtle but important temporal changes. While a number of SSL models have been developed for Alzheimer's prediction, they remain limited by (1) lack of larger, aggregated pre-training datasets \cite{gong,te-ssl,yin}, (2) omission of temporal learning objectives \cite{gong,gryshchuk,yin}, and (3) their inability to handle varying numbers of input images \cite{gong,gryshchuk,te-ssl,yin}. There remains a lack of models that truly exploit temporal self-supervision to improve adaptability and performance across Alzheimer's prediction tasks.

In this paper, we present the first fully adaptable, generalizable self-supervised framework for analyzing 3D longitudinal brain MRI for Alzheimer's prediction tasks.
We adapt three state-of-the-art temporal self-supervised approaches for longitudinal 3D brain MRI analysis and add novel extensions for adaptability to varying numbers of input images and improved spatial features. Specifically, we first implement of temporal order verification, followed by temporal order prediction with a novel approach to support different numbers of input images. Our third model combines the temporal order prediction approach with contrastive learning to improve both spatial and temporal representations. All SSL models are trained using the aggregation of four publicly available datasets (HABS-HD \cite{HABS_HD}, MCSA \cite{MCSA}, NIFD \cite{NIFD}, ADNI \cite{ADNI}) comprising a total of 3,161 patients. We evaluate our models on a subset of held-out test data of CN, MCI and AD patients. Our models are adaptable to taking in one to four images acquired at varying time intervals, showing improved performance with longer image sequences and enabling its application in real-world clinical settings. Further, our model is generalizable across multiple tasks with simple fine-tuning, shown through the classification of CN, MCI, and AD subjects, as well as the detection and future prediction of patients converting from CN to MCI, and from MCI to AD. Through extensive experiments, we find that the temporal order prediction model with contrastive learning consistently has the best performance, and outperforms supervised learning on six of the seven downstream tasks.  
We release all code and models for public use to support reproducibility and enable further development.

\section{Methodology}
\begin{figure}[h!]
    \centering
    \includegraphics[width=0.8\textwidth]{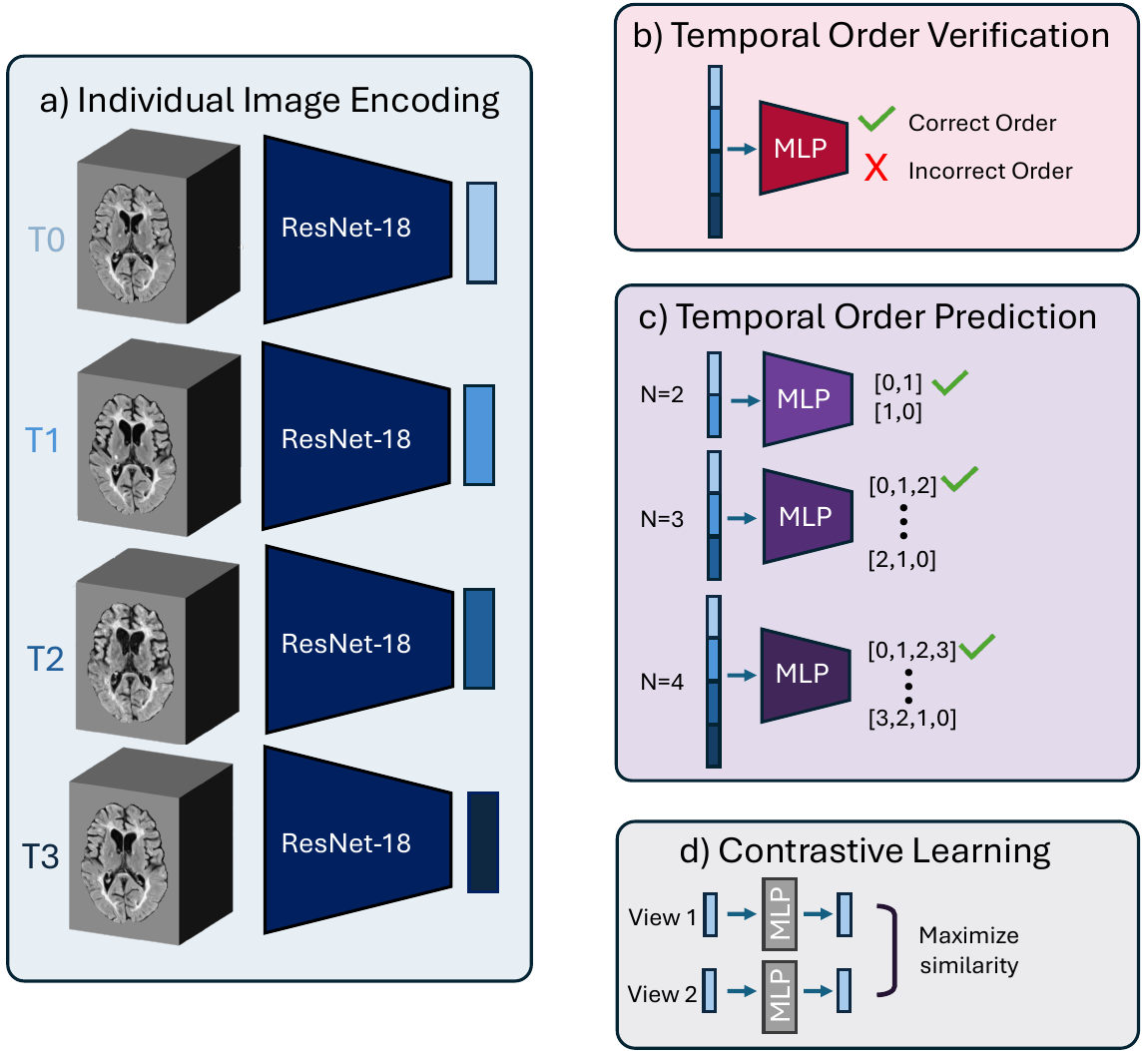}
    \caption{Overview of all three temporal SSL approaches. (a) For each method, up to four 3D brain MRI scans (1-2.5 years apart) are encoded separately by a ResNet-18 backbone. (b) Temporal Order Verification (SSL-TOV) is performed by concatenating image representations and using a multi-layer perceptron (MLP) to predict if the images are in the correct order. If a patient has less than four images, their sequence is padded with zero-filled volumes. (c) Temporal Order Prediction (SSL-TOP) takes the concatenated image representations and predicts the exact permutation of the sequence. Here, three different classifiers are used to predict the permutation for two, three, and four images (no padding is used). (d) The final temporal SSL model (SSL-TOPC) combines temporal order prediction with contrastive learning, where two different views of a sequence are created through augmentations. Images are encoded individually, and the two augmented representations of the first image in a sequence are projected through an MLP. The similarity of the projections is maximized  through the NT-Xent loss.}
    \label{fig:methodology}
\end{figure}

The objective of our work is to develop a temporal self-supervised model that is able to perform challenging detection and prediction tasks that rely on longitudinal information. 
We present an overview of our three temporal self-supervised models for 3D brain MRI analysis, adapted from state-of-the-art SSL methods in natural imaging. These models are designed to correctly predict temporal order of 3D brain MRI sequence; given the similarity in between scans over time, a model must identify key small changes to solve the ordering task. First, we begin with an architecture based on Shuffle and Learn \cite{shuffleandlearn}, a popular method in the context of computer vision videos. During training, we create a binary prediction task to determine if a sequence of 3D brain MRI scans across time is correctly ordered ({\it temporal order verification}), and call this model SSL-TOV. Second, we extend the pre-training task to predicting the exact permutation of the sequence ({\it temporal order prediction} \cite{vcop}), with a novel extension to support classification of two, three, and four MRI input  permutations (SSL-TOP). Third, we combine the temporal order approach with contrastive learning to simultaneously learn temporal and spatial features (SSL-TOPC). All architectures are visualized in Figure \ref{fig:methodology}.

\subsection{SSL-Temporal Order Verification (SSL-TOV)}
The first learning paradigm we examine for longitudinal image analysis is temporal order verification. Given a patient with a sequence of $N$ temporally ordered 3D volumes, $ S = (\mathbf{x}_{t_1}, \mathbf{x}_{t_2}, \dots, \mathbf{x}_{t_N})$, where each $\mathbf{x}_{t_n} \in \mathbb{R}^{C \times D \times H \times W}$ represents a scan acquired at time $t_n$, we randomly permute the sequence and assign a label where \( y = 1 \) if \( t_1 < t_2 < \dots < t_N \), and \( y = 0 \) otherwise. In other words, the sample is labeled positive if all images appear in the correct chronological order, and negative if any image is out of order. After permutation, to account for variability in the number of scans each patient has, we pad sequences to a fixed length $T$ (if $N < T$) by appending zero-filled volumes. Each image in the sequence is then individually encoded by a CNN backbone, $f_\theta(\cdot)$, and the $T$ image representations are concatenated before being passed to multi-layer perceptron classifier, $g_\phi(\cdot)$. The model is trained to minimize the standard binary cross-entropy loss (shown for a single patient): 

\begin{equation}
    \ell_{\text{BCE}} = - \left[ y \log \hat{y} + (1 - y) \log (1 - \hat{y}) \right],
\end{equation}

\noindent where \(y \in \{0,1\}\) is the ground truth label indicating whether the sequence is in correct order, and \(\hat{y}\) is the model's prediction.

\subsection{SSL-Temporal Order Prediction (SSL-TOP)}
\label{sec:tempord}
Following temporal order verification, we consider a more challenging self-supervised task: temporal order prediction. Instead of performing binary classification to predict if the sequence is in order, the model is trained to predict the exact permutation of the image sequence. This formulation increases the task difficulty and encourages the model to focus on specific and subtle differences between all timepoints.

More precisely, for a sequence $S$ of $N$ images, there are $N!$ possible permutations, making the task an $N!$-way classification problem. Since the number of images per patient varies, we develop a novel model modification to enable permutation classification. Specifically, after using the same shared backbone encoder $f_\theta(\cdot)$ as in temporal order verification, the concatenated representations are passed to a permutation classifier corresponding to the sequence length, $g_\phi^{(N)}(\cdot)$. In practice, this means that $g_\phi^{(2)}(\cdot)$ will have $2! = 2$ classes, $g_\phi^{(3)}(\cdot)$ will have $3! = 6$ classes, and so on. The shared backbone of the model ensures that general temporal features are learned, while the permutation classifiers encourage the model to learn specific differences between timepoints in sequences of varying lengths. For a single patient with \(N\) images in their sequence, resulting in \(N!\) classes, the categorical cross-entropy loss is:

\[
\ell^{N}_{\text{CE}} = - \sum_{k=1}^{N!} y_k \log \hat{y}_k,
\]

\noindent where \(y_k\) is the one-hot encoded ground truth permutation and \(\hat{y}_k\) is the predicted probability for class \(k\). This formulation no longer requires image padding, as each sequence length $N$ has its own dedicated classifier. In addition, instead of passing every possible sequence to the classifier each epoch, resulting in the summation of multiple losses $\ell^{N}$ (one loss per classifier head $g_\phi^{(N)}(\cdot)$), we simply sample and train on sequences with the same randomly chosen $N$ for a given epoch (i.e., one epoch focuses on sequences with $N=2$, the next focuses on $N=4$, and so on).

\subsection{SSL-Temporal Order Prediction with Single-Timepoint Contrastive Learning (SSL-TOPC)}

One of the most common and successful forms of self-supervision is contrastive learning. In contrastive learning, two different views of an image are created through augmentations, and the objective is to maximize the similarity between projections of the two augmented representations. To improve spatial feature learning and develop a stronger backbone encoder $f_\theta(\cdot)$, we add single-timepoint contrastive learning to our temporal order prediction model. 

Formally, each temporal brain MRI sequence, $S$, is randomly augmented to create two views, with the same augmentation applied to all images in the sequence:  $S_{i} = a_i(S), S_{j}= a_j(S)$, where $a_i, a_j$ are independent random augmentations (discussed in Section \ref{augs}). After encoding each image in a sequence (without applying any permutation), the representations of the first image in both augmented sequences $\mathbf{h}_{t_1,i} = f_\theta(a_i(x_{t_1})), \mathbf{h}_{t_1,j} = f_\theta(a_j(x_{t_1}))$ are projected through a multi-layer perceptron $\mathbf{z}_{t_1,i} = p_\psi
(\mathbf{h}_{t_1,i}), \mathbf{z}_{t_1,j} = p_\psi
(\mathbf{h}_{t_1,j})$. Only the first image is chosen to reduce the computational burden of applying contrastive learning to all images in a sequence, and we select $t_1$ for simplicity (this likely also improves spatial understanding of baseline images, which are often used in prediction tasks involving one MRI scan). The projections are pushed together to maximize similarity, and pulled apart from all other (first) images in a batch using the NT-Xent loss \cite{simclr} (we remove the $t_1$ subscript for clarity): 

\begin{equation}
    \ell_{i,j} = - \log \frac{\exp(\mathrm{sim}(\mathbf{z}_i, \mathbf{z}_j) / \tau)}{\sum_{k \ne i} \exp(\mathrm{sim}(\mathbf{z}_i, \mathbf{z}_k) / \tau)},
\end{equation}

The total NT-Xent loss for a given patient is calculated using both $z_i$ and $z_j$ as the reference (i.e., the representation used as the comparison in the denominator), summing the loss in both directions:
$\ell_{\text{NT-Xent}} = \ell_{i,j}(\mathbf{z}_i, \mathbf{z}_j) + \ell_{j,i}(\mathbf{z}_j, \mathbf{z}_i)$. Subsequently, the image representations from one of the augmented sequences are randomly permuted, concatenated, and input to the correct permutation prediction head as described in Section \ref{sec:tempord}. The final loss is the sum of the contrastive and classification losses: 

\begin{equation}
    \ell = \ell_{\text{NT-Xent}} + \ell_{\text{CE}}
\end{equation}

Since only the first image in each sequence is used for the contrastive learning loss, we label this as {\it single-timepoint contrastive learning}.

\section{Experiments and Implementation Details}
We pre-train our temporal SSL models on 3D brain MRI aggregated data from four publicly available longitudinal datasets focused on healthy aging and dementia. Each dataset is briefly described below, including our evaluation protocol using ADNI. We evaluate the performance of our models on three specific types of tasks: (1) classification of stable (i.e., constant disease states across input images) CN, MCI, and AD patients using one, two, and three images; (2) detection of conversion from one state to another using two images (e.g., stable CN subjects versus those who converted to MCI); (3) prediction of future conversion to MCI or AD (from CN and MCI, respectively) using a single image. After the dataset and evaluation protocol, we describe our preprocessing and implementation details for the SSL models and supervised baseline. 

\subsection{Datasets}
We use four publicly available datasets for pre-training the SSL models. The \textit{Health and Aging Brain Study – Health Disparities (HABS-HD)} \cite{HABS_HD} and \textit{Mayo Clinic Study of Aging (MCSA)} \cite{MCSA} datasets contain 4,231 and 1,802 patients, respectively, representing CN, MCI and AD patients. The \textit{Neuroimaging in Frontotemporal Dementia (NIFD)} \cite{NIFD} dataset contains 346 healthy controls and patients with frontotemporal or other forms of dementia.

We demonstrate the ability of our model to adapt to different numbers of input images and downstream tasks using the \textit{Alzheimer's Disease Neuroimaging Initiative (ADNI)} \cite{ADNI} dataset, which includes CN, MCI, and AD patients. We divide ADNI into training (2,074 patients), validation (471 patients), and test (483 patients) splits prior to pre-training. The training split is used, in aggregation with the previous three datasets, for SSL pre-training. Given that pre-training focuses on longitudinal image analysis, we omit all patients that only have a single image, creating an SSL pre-training dataset consisting of 3,161 patients. During fine-tuning and training of the supervised baseline, subsets of this training data are selected based on the available labels. The validation and test data are completely held-out from pre-training and are only used for evaluation of downstream tasks.

For consistency across tasks, we maintain these ADNI splits and sample task-specific train/validation/test sets within them. For the classification task, we extract triplets with images that are 1-2.5 years apart with a constant label (e.g., MCI-MCI-MCI). Next, we remove the last image from the triplet to create a pair (e.g., MCI-MCI) and the two last images to create a single timepoint (e.g., MCI). This ensures that the number of patients remains constant across trials, meaning any performance differences are based on input sequence length as opposed to data size. In total, there are 278 patients for training (611 triplets/pairs/images), 80 patients in validation, and 90 patients in test. We perform similar sampling for our conversion detection and prediction tasks: all possible pairs of stable and converting images with a gap of 1-2.5 years are used for training the conversion detection task (e.g., MCI-AD), and the first image in each pair is used for training the future prediction of conversion (e.g., MCI). For CN to MCI conversion, we use 373 patients for training (1,787 scans), 111 patients for validation, and 124 patients for test, and for MCI to AD, we use 516 patients for training (2,483 scans), 157 for validation, and 155 for test.

\subsection{Data Preprocessing}
\label{augs}
All datasets are preprocessed using the same standardized pipeline implemented through TurboPrep\footnote{\url{https://github.com/LemuelPuglisi/turboprep}}
. This publicly available repository has combined common (and fast) 3D brain MRI preprocessing techniques for easy implementation. Briefly, we perform N4 bias-field correction to harmonize and remove inconsistencies from scanners \cite{n4}, skull stripping to remove non-brain regions \cite{skull}, linear registration to align scans to a standard size and shape \cite{registration}, segmentation to identify specific anatomical regions \cite{segmentation}, and intensity normalization for consistent intensities across subjects \cite{normalization}. Images are resampled to a resolution of $1 \times 1 \times 1~\mathrm{mm}^3$ ($193 \times 229 \times 193$ voxels), permuted such that the image depth is first dimension, and cropped to ($150 \times 192 \times 192$). 

During pre-training, we apply augmentations to images across all SSL models using the Medical Open Network for Artificial Intelligence (MONAI \cite{monai}). These augmentations consist of an affine transformation (rotation up to $\pm$ 0.34 rad, translation $\pm$ 15 voxels, scaling in [0.9, 1.3]) applied across all timepoints, followed by independent Gaussian smoothing (sigma in [0.25, 1.5]) and noise (standard deviation in [0.05, 0.09]) in attempt to remove differences between scanners across time. Each augmentation is applied with 0.5 probability. Lastly, we normalize the images by subtracting the overall training set standard deviation. For downstream training or fine-tuning, augmentations include a random spatial crop with minimum size of ($90 \times 115 \times 115$) and resize to ($150 \times 192 \times 192$), random horizontal flip (probability=0.5), random affine transformation with rotation $\pm$0.1 rad, scaling $\pm$ 15\%, and translation $\pm$ 5 voxels (probability=0.7), random intensity shift with offset 0.1 (probability=0.5), and random Gaussian noise added with mean of 0, standard deviation of 0.1 (probability=0.2).

\subsection{Model Details}
For SSL pre-training, we acquire pairs (3,161 patients), triplets (1,114), and quadruplets (445) of images that are 1-2.5 years apart. While additional images may provide increased temporal information, they create increased challenges with missing data, as well as overcomplicated permutation tasks (i.e., five images yields 120 possible classes). The selection of up to four images that are 1-2.5 years apart allows a period of 1-10 years to be analyzed. For patients that have more than a single image/pair/triplet available, we sample a single data input each epoch, to ensure patients are equally represented throughout training.

Each SSL model is implemented with a ResNet-18 backbone and a two-layer multi-layer perceptron classifier of size (512, 256). The projection dimension for the single-timepoint contrastive projection is 128. 
For downstream tasks, we condition all supervised and SSL models on the time gap between scans when training on two or more images. The final layer of the pre-trained SSL classifier (from binary or permutation prediction) is replaced with a linear layer with the desired number of classes. When training with a single image downstream, only the pre-trained backbone is used, and a two-layer multi-perceptron is trained from scratch. For our supervised baseline comparison, we choose a ResNet-18 \cite{resnet}, which is the backbone encoder $f(\cdot)$ for our self-supervised models, to directly evaluate the benefits of SSL pre-training.

\section{Results}
We assess the utility of our pre-training regimes compared against a supervised ResNet-18 model. First, to demonstrate the ability of our model to adapt to varying numbers of input images (and the importance of this ability), we classify stable Alzheimer's patients against stable MCI and CN subjects. Next, using two MRI scans, we detect patients who have converted from one disease state to another. Lastly, using a single MRI scan as input, we predict conversion to another disease state 1-2.5 years in the future, showing the benefit of pre-training on longitudinal objectives to learn generalizable representations with temporal information.

\begin{table}[ht]
\centering
\caption{AUC performance across tasks (higher is better) for the supervised ResNet-18, and SSL-TOV, SSL-TOP, and SSL-TOPC models. Bold indicates best in each row, underline indicates second best. The SSL-TOPC has the highest performance most consistently, outperforming the supervised model on six of the seven tasks. CN: Cognitively Normal; MCI: Mild Cognitive Impairment; AD: Alzheimer's Disease.}

\begin{tabularx}{\textwidth}{lYYYY}
\toprule
Task & \makecell{SL ResNet-18} & 
\makecell{SSL-TOV} & 
\makecell{SSL-TOP} & 
\makecell{SSL-TOPC} \\

\midrule
\multicolumn{5}{l}{\textbf{Classification of Stable CN, MCI, and AD Patients}} \\
1 MRI Scan & 0.673$\pm$0.041 & \underline{0.683$\pm$0.037} & 0.680$\pm$0.048 & \textbf{0.712$\pm$0.057} \\
2 MRI Scans & \underline{0.767$\pm$0.015} & 0.748$\pm$0.019 & 0.745$\pm$0.021 & \textbf{0.789$\pm$0.028} \\
3 MRI Scans & 0.780$\pm$0.020 & \underline{0.848$\pm$0.011} & \textbf{0.872$\pm$0.030} & 0.838$\pm$0.004 \\
\midrule
\multicolumn{5}{l}{\textbf{Conversion Detection with Two Images}}\\
CN $\rightarrow$ MCI (2 MRI)   & \underline{0.639$\pm$0.042} & 0.634$\pm$0.037 & 0.610$\pm$0.128 & \textbf{0.650$\pm$0.057} \\
MCI $\rightarrow$ AD (2 MRI)    & \underline{0.769$\pm$0.034} & 0.722$\pm$0.028 & \textbf{0.778$\pm$0.021} & 0.719$\pm$ 0.024\\
\midrule
\multicolumn{5}{l}{\textbf{Future Conversion Prediction with a Single Image}}\\
CN $\rightarrow$ MCI (1 MRI)    & \underline{0.680$\pm$0.018} & 0.653$\pm$0.056 & 0.642$\pm$0.038 & \textbf{0.697$\pm$0.053} \\
MCI $\rightarrow$ AD (1 MRI)   & 0.633$\pm$0.043 & \underline{0.686$\pm$0.023} & \textbf{0.693$\pm$0.021} & 0.667$\pm$0.045 \\
\bottomrule
\end{tabularx}
\label{tab:results}
\end{table}

Table \ref{tab:results} shows AUC performance for seven different Alzheimer's prediction tasks. We repeat each downstream task three times and report the mean and standard deviation. Overall, the SSL-TOPC model shows superior performance for six out of the seven tasks compared to supervised learning. Importantly, we notice that this model particularly excels when training on a single image. For example, when classifying stable CN, MCI, and AD patients, and predicting future conversion of CN to MCI and MCI to AD, the SSL-TOPC has a 4\%, 1\%, and 3\% increase over supervised baselines (respectively), and higher performance than the other SSL models for two of these tasks. This highlights the benefit of extracting strong spatial features through contrastive learning, particularly when temporal information is unavailable in downstream applications. The SSL-TOV and SSL-TOP models do not outperform the supervised ResNet-18 as often as SSL-TOPC, further demonstrating the importance of combining temporal and spatial SSL learning objectives.
\subsection{The Effect of Multiple Timepoints}
The first three rows of Table \ref{tab:results} show the performance of classifying stable CN, MCI, and AD patients, using one, two, or three input images. All three SSL models have a significant increase in performance over the supervised baseline when fine-tuning on three images. This indicates that these models have learned temporal information useful for longer-term classification tasks. In general, all models perform better as the number of input scans increases, highlighting the importance of developing models that can adapt to varying numbers of input images to leverage all available information for improved prediction. 

\subsection{Conversion Detection}
The next task performed is conversion detection (rows 4 and 5 in Table \ref{tab:results}), where two images are used to classify if conversion has occurred. Interestingly, the supervised ResNet-18 performs much better than the SSL-TOPC model, and almost as well as the SSL-TOP model at detecting the conversion from MCI to AD. Alzheimer's disease has significant visible brain changes such as increased atrophy, which are likely easier to detect than the subtle changes in the CN-to-MCI task. However, for the more difficult task of predicting the change of CN to MCI, the SSL-TOPC model excels, displaying the advantage of spatiotemporal pre-training. 

\subsection{Future Conversion Prediction}
The final and most difficult tasks involve predicting future conversion (within the next 1-2.5 years) from a single MRI scan (Table \ref{tab:results}, rows 6 and 7). As mentioned previously, the SSL model with contrastive learning achieves strong results in predicting future conversion from CN to MCI, showing the importance of learning robust spatial features for single-image prediction tasks. All three SSL models perform well for the prediction of MCI to AD. The higher performance can likely be attributed to the temporal pre-training objective, which encourages the model to understand changes over time and therefore strongly contributes to improving prediction of future disease conversion.

\section{Conclusion}
In this work, we adapt three different state-of-the-art temporal self-supervised pre-training approaches for Alzheimer's prediction tasks and include novel extensions to address the challenges of real-world medical images. We demonstrate that pre-training with increased data and temporal learning objectives can significantly improve the downstream performance, with notable improvement when combining temporal order prediction with contrastive learning. Importantly, the SSL-TOPC model trained using both temporal order prediction and contrastive learning achieves superior performance over supervised learning for six of the seven tasks, and is adaptable across different numbers of input images with varying time intervals, as well as generalizable across numerous downstream tasks. We release all code and pre-trained models publicly, with the intention of providing a generalizable and adaptable model tailored to longitudinal and/or Alzheimer's prediction tasks.

\begin{credits}
\subsubsection{\ackname}
This work was supported by the Natural Sciences and Engineering Research Council of Canada, Fonds de Recherche du Quebec: Nature et Technologies, the Canadian Institute for Advanced Research (CIFAR) Artificial Intelligence Chairs program, Calcul Quebec, the Digital Research Alliance of Canada, the Vadasz Scholar McGill Engineering Doctoral Award, Mila - Quebec AI Institute, the International Progressive MS Alliance and the MS Society of Canada.

We also acknowledge all datasets used in this work. HABS-HD: Research reported on this publication was supported by the National Institute on Aging of the National Institutes of Health under Award Numbers R01AG054073, R01AG058533, P41EB015922 and U19AG078109.
MCSA: The data contained in this analysis were obtained under research grant from the National Institutes of Health to the Mayo Clinic Study of Aging (U01 AG006786, Ronald Petersen, PI).
NIFD: Data collection and sharing for this project was funded by the Frontotemporal Lobar Degeneration Neuroimaging Initiative (National Institutes of Health Grant R01 AG032306). 
ADNI: Data collection and sharing for the Alzheimer's Disease Neuroimaging Initiative (ADNI) is funded by the National Institute on Aging (National Institutes of Health Grant U19AG024904).

\subsubsection{\discintname} The authors have no competing interests to declare that are relevant to the content of this article.

\end{credits}

\bibliographystyle{splncs04}
\bibliography{references}

\begin{thebibliography}{10}
\providecommand{\url}[1]{\texttt{#1}}
\providecommand{\urlprefix}{URL }
\providecommand{\doi}[1]{https://doi.org/#1}

\bibitem{registration}
Avants, B., Epstein, C., Grossman, M., Gee, J.: Symmetric diffeomorphic image registration with cross-correlation: Evaluating automated labeling of elderly and neurodegenerative brain. Medical Image Analysis  \textbf{12}(1),  26--41 (2008). \doi{https://doi.org/10.1016/j.media.2007.06.004}, special Issue on The Third International Workshop on Biomedical Image Registration – WBIR 2006

\bibitem{segmentation}
Billot, B., Greve, D.N., Puonti, O., Thielscher, A., {Van Leemput}, K., Fischl, B., et~al.: Synthseg: Segmentation of brain mri scans of any contrast and resolution without retraining. Medical Image Analysis  \textbf{86},  102789 (2023). \doi{https://doi.org/10.1016/j.media.2023.102789}

\bibitem{monai}
Cardoso, M.J., Li, W., Brown, R., Ma, N., Kerfoot, E., Wang, Y., et~al.: Monai: An open-source framework for deep learning in healthcare. arXiv:2211.02701  (2022)

\bibitem{simclr}
Chen, T., Kornblith, S., Norouzi, M., Hinton, G.: A simple framework for contrastive learning of visual representations. In: Proceedings of the 37th International Conference on Machine Learning. ICML'20 (2020)

\bibitem{ericsson}
Ericsson, L., Gouk, H., Loy, C.C., Hospedales, T.M.: Self-supervised representation learning: Introduction, advances, and challenges. IEEE Signal Processing Magazine  \textbf{39}(3),  42--62 (2022). \doi{10.1109/MSP.2021.3134634}

\bibitem{sina}
Fathi, S., Ahmadi, M., Dehnad, A.: Early diagnosis of alzheimer's disease based on deep learning: A systematic review. Computers in Biology and Medicine  \textbf{146},  105634 (2022). \doi{https://doi.org/10.1016/j.compbiomed.2022.105634}

\bibitem{tory}
Frizzell, T.O., Glashutter, M., Liu, C.C., Zeng, A., Pan, D., Hajra, S.G., et~al.: Artificial intelligence in brain mri analysis of alzheimer’s disease over the past 12 years: A systematic review. Ageing Research Reviews  \textbf{77},  101614 (2022). \doi{https://doi.org/10.1016/j.arr.2022.101614}

\bibitem{gong}
Gong, H., Wang, Z., Huang, S., Wang, J.: A simple self-supervised learning framework with patch-based data augmentation in diagnosis of alzheimer’s disease. Biomedical Signal Processing and Control  \textbf{96},  106572 (2024). \doi{https://doi.org/10.1016/j.bspc.2024.106572}

\bibitem{gryshchuk}
Gryshchuk, V., Singh, D., Teipel, S., Dyrba, M.: Contrastive self-supervised learning for neurodegenerative disorder classification. Frontiers in Neuroinformatics  \textbf{19} (2025)

\bibitem{resnet}
He, K., Zhang, X., Ren, S., Sun, J.: Deep residual learning for image recognition. Eprint \href{https://arxiv.org/abs/1512.03385}{arXiv:1512.03385} (2015)

\bibitem{skull}
Hoopes, A., Mora, J.S., Dalca, A.V., Fischl, B., Hoffmann, M.: Synthstrip: skull-stripping for any brain image. NeuroImage  \textbf{260},  119474 (2022). \doi{https://doi.org/10.1016/j.neuroimage.2022.119474}

\bibitem{Taeho}
Jo, T., Nho, K., Saykin, A.J.: Deep learning in alzheimer's disease: Diagnostic classification and prognostic prediction using neuroimaging data. Frontiers in Aging Neuroscience  \textbf{11} (2019)

\bibitem{admci}
Kelley, B.J., Petersen, R.C.: Alzheimer's disease and mild cognitive impairment. Neurologic Clinics  \textbf{25}(3),  577--609 (2007). \doi{https://doi.org/10.1016/j.ncl.2007.03.008}, dementia

\bibitem{shuffleandlearn}
Misra, I., Zitnick, C.L., Hebert, M.: Shuffle and learn: Unsupervised learning using temporal order verification. In: Leibe, B., Matas, J., Sebe, N., Welling, M. (eds.) Computer Vision - {ECCV} 2016 - 14th European Conference, Amsterdam, The Netherlands, October 11-14, 2016, Proceedings, Part {I}. Lecture Notes in Computer Science, vol.~9905, pp. 527--544. Springer (2016). \doi{10.1007/978-3-319-46448-0\_32}

\bibitem{HABS_HD}
O'Bryant, S.E., Johnson, L.A., Barber, R.C., Braskie, M.N., Christian, B., Hall, J.R., et~al.: The health \& aging brain among latino elders (hable) study methods and participant characteristics. Alzheimer's \& Dementia: Diagnosis, Assessment \& Disease Monitoring  \textbf{13}(1),  e12202 (2021)

\bibitem{ADNI}
Petersen, R.C., Aisen, P.S., Beckett, L.A., Donohue, M.C., Gamst, A.C., Harvey, D.J., et~al.: Alzheimer's disease neuroimaging initiative (adni) clinical characterization. Neurology  \textbf{74}(3),  201--209 (2010)

\bibitem{MCSA}
Roberts, R.O., Geda, Y.E., Knopman, D.S., Cha, R.H., Pankratz, V.S., Boeve, B.F., et~al.: The mayo clinic study of aging: design and sampling, participation, baseline measures and sample characteristics. Neuroepidemiology  \textbf{30}(1),  58--69 (2008)

\bibitem{NIFD}
Rosen, H.J., the FTLDNI Study~Group: Frontotemporal lobar degeneration neuroimaging initiative (ftldni). Funded by the National Institute on Aging; coordinated by UCSF and collaborating sites in North America (2010), \url{http://memory.ucsf.edu/research/studies/nifd}, data used in this study were obtained from the FTLDNI database.

\bibitem{normalization}
Shinohara, R.T., Sweeney, E.M., Goldsmith, J., Shiee, N., Mateen, F.J., Calabresi, P.A., et~al.: Statistical normalization techniques for magnetic resonance imaging. NeuroImage: Clinical  \textbf{6},  9--19 (2014). \doi{https://doi.org/10.1016/j.nicl.2014.08.008}

\bibitem{te-ssl}
Thrasher, J., Devkota, A., Tafti, A.P., Bhattarai, B., Gyawali, P.: Te-ssl: Time and event-aware self supervised learning for alzheimer's disease progression analysis. In: Linguraru, M.G., Dou, Q., Feragen, A., Giannarou, S., Glocker, B., Lekadir, K., Schnabel, J.A. (eds.) Medical Image Computing and Computer Assisted Intervention -- MICCAI 2024. pp. 324--333. Springer Nature Switzerland, Cham (2024)

\bibitem{n4}
Tustison, N.J., Avants, B.B., Cook, P.A., Zheng, Y., Egan, A., Yushkevich, P.A., Gee, J.C.: N4itk: Improved n3 bias correction. IEEE Transactions on Medical Imaging  \textbf{29}(6),  1310--1320 (2010). \doi{10.1109/TMI.2010.2046908}

\bibitem{valizadeh}
Valizadeh, G., Elahi, R., Hasankhani, Z., Rad, H.S., Shalbaf, A.: Deep learning approaches for early prediction of conversion from mci to ad using mri and clinical data: A systematic review. Archives of Computational Methods in Engineering  \textbf{32},  1229--1298 (2025)

\bibitem{vcop}
Xu, D., Xiao, J., Zhao, Z., Shao, J., Xie, D., Zhuang, Y.: Self-supervised spatiotemporal learning via video clip order prediction. In: Computer Vision and Pattern Recognition (CVPR) (2019)

\bibitem{yin}
Yin, Y., Jin, W., Bai, J., Liu, R., Zhen, H.: Smil-deit:multiple instance learning and self-supervised vision transformer network for early alzheimer's disease classification. In: 2022 International Joint Conference on Neural Networks (IJCNN). pp.~1--6 (2022). \doi{10.1109/IJCNN55064.2022.9892524}

\end{thebibliography}

\end{document}